\documentclass[11pt,a4paper]{article}
\usepackage[hyperref]{emnlp2020}
\usepackage{times}
\usepackage{latexsym}
\usepackage{amsmath}
\usepackage{multirow} 
\usepackage{booktabs}

\usepackage[utf8]{inputenc}
\usepackage{graphicx}
\usepackage{microtype}
\usepackage{url}
\aclfinalcopy % Uncomment this line for the final submission
\title{ Volctrans Parallel Corpus Filtering System for WMT 2020}

\author{Runxin Xu$^{1,3}$, Zhuo Zhi$^{2,3}$, Jun Cao$^3$, Mingxuan Wang$^3$, Lei Li$^3$ \\
$^1$ Institute of Computational Linguistic, Peking University \\
$^2$ Department of Computer Science and Engineering, University of California San Diego \\
$^3$ ByteDance AI Lab, Shanghai, China \\
\texttt{runxinxu@gmail.com} \\
\texttt{zzhi@ucsd.edu} \\
\texttt{\{caojun.sh, wangmingxuan.89, lileilab\}@bytedance.com}
}
\date{}

\begin{document}
\maketitle
\begin{abstract}
In this paper, we describe our submissions to the WMT20 shared task on parallel corpus filtering and alignment for low-resource conditions.
The task requires the participants to align potential parallel sentence pairs out of the given document pairs, and score them so that low-quality pairs can be filtered.
Our system, Volctrans, is made of two modules, i.e., a mining module and a scoring module.
Based on the word alignment model, the mining module adopts an iterative mining strategy to extract latent parallel sentences.
In the scoring module, an XLM-based scorer provides scores, followed by reranking mechanisms and ensemble.
Our submissions outperform the baseline by 3.x/2.x and 2.x/2.x for km-en and ps-en on From Scratch/Fine-Tune conditions, which is the highest among all submissions.

\end{abstract}

\section{Introduction}

With the rapid development of machine translation, especially \textbf{N}eural \textbf{M}achine \textbf{Translation} (NMT) \citep{DBLP:conf/nips/VaswaniSPUJGKP17, ott-etal-2018-scaling,DBLP:conf/iclr/ZhuXWHQZLL20}, parallel corpus in high quality and large quantity is in urgent demand.
These parallel corpora can be used to train and build robust machine translation models. 
However, for some language pairs on low-resource conditions, few parallel resources are available. 
Since it is much easier to obtain quantities of monolingual data, it may help if we can extract parallel sentences from monolingual data through alignment and filtering.

The WMT19 shared task on parallel corpus filtering for low-resource conditions \citep{koehn-etal-2019-findings} provides
noisy parallel corpora in Sinhala-English and Nepali-Englis crawled from the web. Participants are asked to score sentence pairs so that low-quality sentences are filtered.
In this year, the WMT20 shared task on parallel Corpus filtering and alignment for low-resource conditions is very similar, except that the language pairs become Khmer-English and Pashto-English, and the provided raw data are documents in pair, which require sentence-level alignment.
Besides, no data in similar languages are provided for this year.

The participants are required to align sentences within documents in different languages and provide a score for each sentence pair.
To evaluate the quality of the extracted sentence pairs, they are subsampled to $5$ million English words and used to train a neural machine translation model. 
Finally, the BLEU score of the machine translation system is used to reflect the quality of the sentence pairs.

In this paper, we propose the Volctrans filtering system, which consists of a \textbf{mining module} and a \textbf{scoring module}. 
First, the mining module extracts and aligns potential parallel sentence pairs within documents in different languages. 
In particular, we introduce an iterative mining strategy to boost mining performance.
We keep adding newly aligned high-quality parallel sentences to train the word alignment model, which is essential for the mining module.
Second, the scoring module is based on XLM \citep{DBLP:conf/nips/ConneauL19}, and responsible for providing scores for each sentence pair.
Several reranking mechanisms are also used in this module.
We conduct experiments to tune the hyper-parameters for the best filtering performance, and four systems are ensembled to achieve the final results.

\section{System Architecture}

%In this section, we describe our proposed system. 
%First, we briefly introduce the provided data.
%Next, the mining module and scoring module are introduced respectively. 

\subsection{Data Introduction}
In detail, as is shown in Table~\ref{tab:data}, the WMT20 shared task provides:
\begin{itemize}
\item Document pairs, including $391,250$ Khmer-English and $45,312$ Pashto-English document pairs;
\item Sentence-aligned corpora extracted from the above document pairs, using Hunalign \citep{DBLP:conf/lrec/2008} and LASER \cite{DBLP:journals/tacl/ArtetxeS19}, including $4,169,574$ Khmer-English and $1,022,883$ Pashto-English sentence pairs;
\item Parallel data which can be used to build filtering and alignment system, including $290,051$ Khmer-English and $123,198$ Pashto-English parallel sentences;
\item Monolingual data, including approximately $1.9$ billion English, $14.0$ million Khmer, and $6.6$ million Pashto sentences. 
\end{itemize}

\begin{table}[htbp]
\centering
\caption{Statistics of Provided Data Scale}
\label{tab:data}
\begin{tabular}{ccc}
 & en-km & en-ps\\
\hline
Document Pairs & 391K & 45K \\
Extracted Sentence Pairs & 4.2M & 1.0M \\
Parallel Sentences & 290K & 123K \\
\hline
\end{tabular}
\end{table}

\subsection{Mining Module}

\begin{figure}[htp]
    \centering
    \includegraphics[width=0.45\textwidth,keepaspectratio]{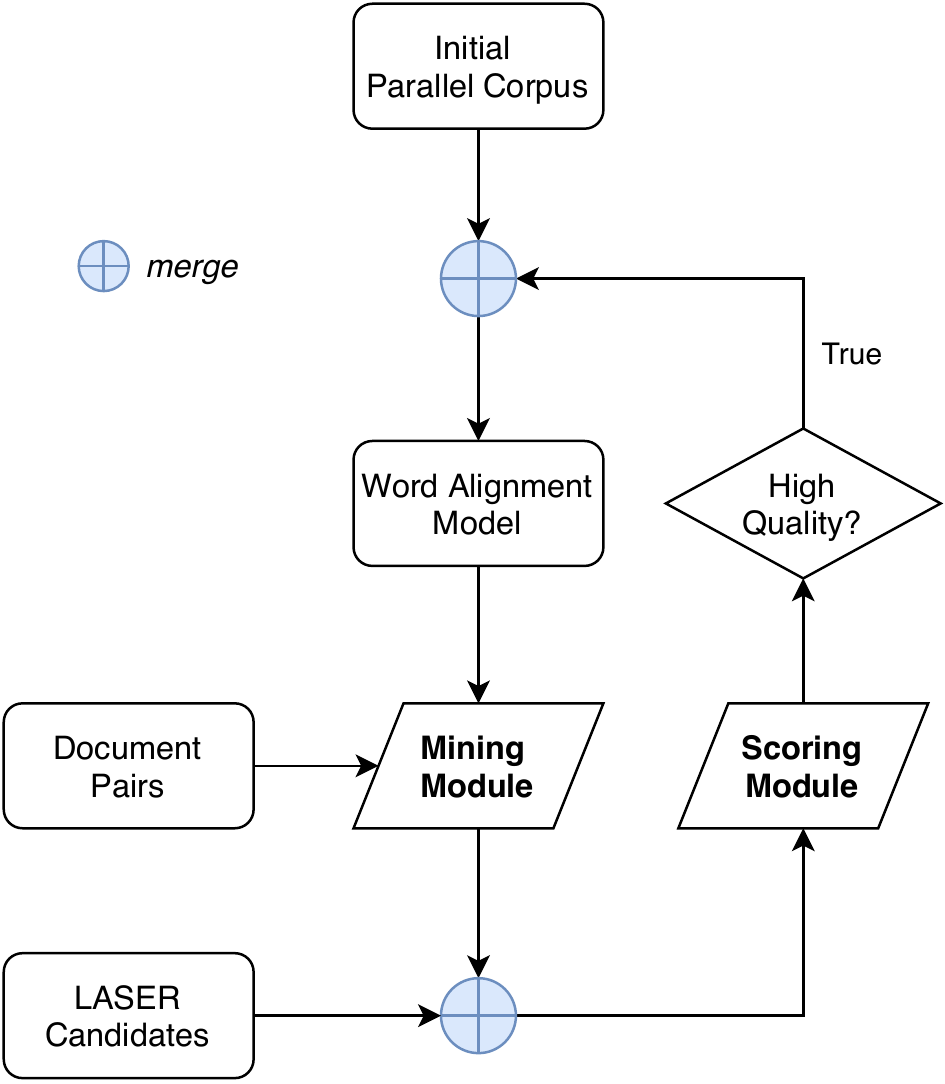}
    \caption{Mining Procedure}
    \label{fig:mining}
\end{figure}

Besides the given aligned sentences, we believe that there are still more potential parallel sentences that can be mined. Thus we choose to extract our own set of sentence pairs from the provided document pairs, and design a mining module aiming to gather as many parallel sentence candidates as possible. We then elaborate on our mining procedure and mining module, shown in Figure~\ref{fig:mining}, in detail.

%After careful analysis of the provided data, we found most of the document pairs come from identical domain sites for each pair, and usually differ on the language code part. Compared with the monolingual data from Wikipedia or CommonCrawl, this part of data are likely to contain potential parallel sentences with higher confidence, but cost less computation effort. Therefore, we focus on aligning sentences from these document pairs. We then elaborate on our mining procedure and mining module, shown in Figure~\ref{fig:mining}, in detail.

%The procedure is taken as following steps:
%\begin{enumerate}
%\item \textbf{Word alignment}. 
\subsubsection{Word Alignment}
\label{word-alignment}
We trained the word alignment model on the provided clean parallel corpus by using the \textit{fast-align} toolkit \citep{dyer-etal-2013-simple}, and get the forward and reverse word translation probability tables. It's worth mentioning that both of Pashto and Khmer corpus are tokenized before word alignment model training for accuracy consideration. We separate Pashto words by moses tokenizer\footnote{\url{https://github.com/moses-smt/mosesdecoder/blob/master/scripts/tokenizer/tokenizer.perl}}. For Khmer, we use the character (\textbackslash u200B in Unicode) as separator when it's available and otherwise use a dictionary-based tokenizer by maximizing the word sequence probability. 

%\item \textbf{Mining parallel sentences}. 
\subsubsection{Mining Parallel Sentences}
This step is operated by our mining module. With the bilingual word translation probability tables, the mining module evaluates the translation quality of bilingual sentence pairs by YiSi-2 \citep{lo-2019-yisi}, which involves both lexical weight and lexical similarity. The Document pairs are first segmented on each language side using Polyglot \footnote{\url{https://github.com/aboSamoor/polyglot}}. This initial segmentation is represented as:
\begin{align}
    e &= \overline{e}_1 \overline{e}_2 \cdots \overline{e}_a = \overline{e}_1^a \\
    f &= \overline{f}_1 \overline{f}_2 \cdots \overline{f}_b = \overline{f}_1^b
\end{align}
where $\overline{e}_k$ ($\overline{f}_k$) is a segment of consecutive words of document $e$ ($f$). Then we compute the sentence similarity (translation quality) by iteration from the initial segment ($\overline{e}_{1}$, $\overline{f}_{1}$). If the similarity reaches the preset threshold for ($\overline{e}_{i}$, $\overline{f}_{j}$), we pick the segment pair as parallel sentence candidate, and continue the computation from ($\overline{e}_{i+1}$, $\overline{f}_{j+1}$).

%\item \textbf{Sentence similarity}. With the bilingual word translation probability tables, we evaluate the translation quality of bilingual sentence pairs by YiSi-2 \citep{lo-2019-yisi}, which involves both lexical weight and lexical similarity.

%\item \textbf{Document segmentation}. Document pairs are then segmented on each language side using Polyglot \footnote{\url{https://github.com/aboSamoor/polyglot}}. This initial segmentation is represented as:
%\begin{align}
%    e &= \overline{e}_1 \overline{e}_2 \cdots \overline{e}_a  \\
%    f &= \overline{f}_1 \overline{f}_2 \cdots \overline{f}_b 
%\end{align}
%We compute the sentence similarity by iteration from the initial segment ($\overline{e}_{1}$, $\overline{f}_{1}$). If the similarity reaches the preset threshold for ($\overline{e}_{i}$, $\overline{f}_{j}$), we pick the segment pair as parallel sentence candidate, and continue the computation from ($\overline{e}_{i+1}$, $\overline{f}_{j+1}$).

We notice that the inconsistency of segmentation in the document pairs can 
%have a bad effect on sentence alignment. Sentences segmented without utilizing their cross-lingual alignment information in the document pairs can 
lead to the results: a sentence in one language contains information only part of a sentence in the other language, or two sentences (in different languages) both contain part of their information in common. These resulting sentence pairs may have low similarity scores.

In order to alleviate this problem, we also incorporate a parallel segmentation method in our mining module. %Assuming that the document pairs are aligned correctly, we aim to find appropriate split points using alignment information to produce satisfied aligned segments. 
We follow the basic idea proposed in \citep{nevado-etal-2003-parallel} where the parallel segmentation finding problem is treated as an optimization problem and a dynamic programming scheme is used to search for the best segmentation. We then briefly introduce our method. \footnote{More details can be found in \citep{nevado-etal-2003-parallel}}

%Given an aligned document pair $(e,f)$, we first obtain monolingual initial segmentation using Polyglot. This initial segmentation is represented as:
%\begin{align}
%    e &= \overline{e}_1 \overline{e}_2 \cdots \overline{e}_a = \overline{e}_1^a \\
%    f &= \overline{f}_1 \overline{f}_2 \cdots \overline{f}_b = \overline{f}_1^b
%\end{align}
%where $\overline{e}_k$ ($\overline{f}_k$) is a segment of consecutive words of document $e$ ($f$) and $\overline{e}_{k_1}^{k_2}$ ($\overline{f}_{k_1}^{k_2}$) are consecutive segments $\overline{e}_{k_1} \overline{e}_{k_1+1} \cdots \overline{e}_{k_2}$ ($\overline{f}_{k_1} \overline{f}_{k_1+1} \cdots \overline{f}_{k_2}$).

After obtaining the monolingual initial segmentation $(\overline{e}_1^a,\overline{f}_1^b)$, a parallel segmentation is represented as:
\begin{equation}
\begin{split}
    s \equiv ([\overline{e}_{1}^{k_1}, \overline{f}_{1}^{j_1}], [\overline{e}_{k_1+1}^{k_2}, \overline{f}_{j_1+1}^{j_2}], \\
    \cdots, [\overline{e}_{k_{|s|-1}+1}^{k_|s|}, \overline{f}_{j_{|s|-1}+1}^{j_|s|}])
\end{split}
\end{equation}
where $|s|$ is the number of segments for the parallel segmentation $s$ and $\overline{e}_{i_1}^{i_2}$ ($\overline{f}_{i_1}^{i_2}$) are consecutive segments $\overline{e}_{i_1} \overline{e}_{i_1+1} \cdots \overline{e}_{i_2}$ ($\overline{f}_{i_1} \overline{f}_{i_1+1} \cdots \overline{f}_{i_2}$). 
%we have $k_{|s|} = a$ and $j_{|s|} = b$. 
In this setting, all initial segments will be included in the parallel segmentation, and the order of the initial segments cannot be inverse. Therefore, the alignment is monotone.

%In order to find the best parallel segmentation, we then turn to translation model to construct the objective function. Let's denote the probability of translating a document $e$ into $f$ as $P(f|e)$, then we can further extend this equation as:
%\begin{equation}
%\begin{split}
%    P(f|e) = P(\overline{f}_1^b|\overline{e}_1^a) = \sum\limits_{S \in s} P(\overline{f}_1^b,S|\overline{e}_1^a) \\
    %= \sum\limits_{S \in s} P(S|\overline{e}_1^a) \prod\limits_{n=1}^{|s|} %P(\overline{f}_{j_{n-1}+1}^{j_n}|\overline{e}_{k_{n-1}+1}^{k_n})
%\end{split}
%\end{equation}

%Since we want to search for the best segmentation, we define the most probable segmentation probability as:
%\begin{equation}
%    \hat{P}(\overline{f}_1^b|\overline{e}_1^a) = \max\limits_{S \in s} P(S|\overline{e}_1^a) \prod\limits_{n=1}^{|s|} P(\overline{f}_{j_{n-1}+1}^{j_n}|\overline{e}_{k_{n-1}+1}^{k_n})
%\end{equation}

%Considering $P(S|\overline{e}_1^a)$ has a uniform distribution ($P(S|\overline{e}_1^a) = C$), we can write our objective function as:
Then we search for the best parallel segmentation using the objective function:
\begin{equation}
    \max\limits_{S \in s} C \cdot \prod\limits_{n=1}^{|s|} P(\overline{f}_{j_{n-1}+1}^{j_n}|\overline{e}_{k_{n-1}+1}^{k_n})
\end{equation}
where $P(\overline{f}_{j_{n-1}+1}^{j_n}|\overline{e}_{k_{n-1}+1}^{k_n})$ is the translation quality of the pair $(\overline{f}_{j_{n-1}+1}^{j_n},\overline{e}_{k_{n-1}+1}^{k_n})$. 

Next, we use a dynamic programming algorithm to compute the best segmentation $S$ where we have a restriction that no more than 3 initial segments can be joined.

Finally, this set of parallel segments are combined with the set of the extracted initial segment pairs through global deduplication to serve as the output of our mining module.

It is worth noting that the sequence can be very long in the process of translation quality computation because several segments can be joined together. Therefore, while computing the similarity of a segment pair, our method based on word translation probability tables can be more time-efficient than LASER, as LASER is based on sentence embeddings and can be very slow when its LSTM encoder is fed with long sequence. Thus we do not consider using LASER to compute translation quality in our mining procedure.

%\item \textbf{Scoring for boosting}. 
\subsubsection{Iterative Mining Strategy}
% With the given sentence pairs by LASER and our mined parallel sentence candidates, we further utilize a scoring module which will be introduced in section \ref{scoring module} to refine the outcome with high quality. 
The quantity and quality of the mined data are basically dependent on the word alignment model.
Besides, more high-quality parallel corpus is used, the word alignment model would be more accurate and robust.
Therefore, we propose an iterative mining strategy.
For the first time, all the provided parallel data by the task are used to train the word alignment model.
But we keep mining data for several times.
We iteratively add new high-quality sentence pairs to the parallel corpus and train the word alignment model again to improve the word translation probability tables, thus boosting the mining cycle.

\subsection{Scoring Module} \label{scoring module}
The scoring module consists of three parts. 
First, we make use of both the parallel and monolingual data to train an XLM-based classifier to score each sentence pair. 
Secondly, different reranking mechanisms are used to adjust the scores. 
Finally, we ensemble four different models to improve the performance of our systems.

\subsubsection{XLM-based Scorer}
Recently, pre-trained transformer-based models play an important role in a variety of NLP tasks, such as question answering, relation extraction, etc.. 
Pre-trained models are often trained from scratch with self-supervised objective, and then fine-tuned to adapt to the downstream tasks.
In our system, we choose the XLM \citep{DBLP:conf/nips/ConneauL19} as our main model.
The reason are as follows:
a) Similar to BERT \citep{devlin-etal-2019-bert,yang2019towards}, XLM has \textbf{M}asked \textbf{L}anguage \textbf{M}odel (MLM) objective, which enables us to make the most use of the provided monolingual corpora;
b) XLM also has \textbf{T}ranslation \textbf{L}anguage \textbf{M}odel (TLM) objective. Taking two parallel sentences as input, it predicts the randomly masked tokens. In this way, cross-lingual features can be captured;
c) With a large amount of training corpus in different languages, XLM can provide powerful cross-lingual representation for downstream tasks, which is very suitable for parallel corpus filtering situations.

We follow the instructions \footnote{\url{https://github.com/facebookresearch/XLM}} to prepare the training data and train the XLM model.
In detail, we use Moses tokenizer to tokenize the text \footnote{We do not use the character-/dictionary- based method introduced in Section~\ref{word-alignment} to tokenize Khmer here. Performance may be improved with that method, but we have run out of time,  unfortunately.}, and 
fastbpe \footnote{\url{https://github.com/glample/fastBPE}} to learn and apply Byte-Pair Encoding. 
We use $50$K BPE codes on the concatenation of all the training data.
After applying BPE codes to the training data, we obtain a large vocabulary containing around $100$K tokens. 
Therefore, we only keep the top frequent $100,000$ tokens to form the vocabulary and train the XLM.

We use monolingual data in MLM objective and parallel data in TLM objective. 
In detail, the monolingual data we use are as follows:
\begin{itemize}
\item Khmer: all the $14$M provided sentences.
\item Pashto: all the $6.6$M provided sentences.
\item English: because the number of english monolingual sentences are so large, we subsample $25$M sentences to keep a balance.
\end{itemize}

All the available parallel sentence pairs ($29$K en-km and $12$K en-ps) are used in TLM objective.
For each objective, we hold out $5$K sentences or sentence pairs for validation and $5$K for the test.

We pre-train the XLM using two different settings on $8$ Tesla-V100 GPU:
a) \textbf{Standard}: The embedding size is $1024$, with $12$ layers and $64$ batch size.
b) \textbf{Small}: The embedding size is $512$, with $6$ layers and $32$ batch size.
The other values of hyperparameters are all set to the default values. The two pre-trained XLM model is then fine-tuned in downstream task and further ensembled.

To score sentence pairs according to their parallelism, classification models are usually used \citep{xu-koehn-2017-zipporah, bernier-colborne-lo-2019-nrc}. 
In the training phrase, it is formulated as a binary classification problem, whether the sentence pair is semantically similar to each other or not. 
In the inference phrase, the probability of the positive class is considered as the score of the sentence pair.
Therefore, we use the provided parallel sentence pairs as the positive instances, and construct negative instances taking advantage of such positive instances similar to \citet{bernier-colborne-lo-2019-nrc}.
Specifically, we generate negative examples in the following ways:
\begin{itemize}
    \item Shuffle the sentences in source language and target language respectively, and randomly align them.
    \item Randomly truncate the length of the source sentences or/and target sentences to $3$.
    \item Randomly shuffle the order of the source sentences or/and target sentences.
    \item Simply swap the source and target sentences. Or replace the source/target sentences with target/source sentences, such that the two sentences are exactly the same.
\end{itemize}

We only add a linear or convolutional layer on top of the pre-trained XLM model and predict through a sigmoid function. 
The input of the model is the concatenation of a sentence pair, separated by one [SEP] token. 
Besides, to tackle the problem that some sentences may be too long, we simply truncate each sentence such that the maximum length of sentence is $128$. 
The dropout rate is set to $0.5$.

\subsubsection{Reranking}
We apply some reranking mechanisms in order to compensate for the latent bias in the XLM-based scorer, and aim to boost the quality of the whole corpus rather than each sentence pair independently.

The first reranking mechanism is based on \textbf{language identification}. 
For some sentences, they may include many tokens that do not belong to the corresponding language, and therefore damage the performance of the machine translation system.
This phenomenon is rather common in Khmer-English corpus in particular.
We utilize \textit{pycld2} tools \footnote{\url{https://github.com/aboSamoor/pycld2}} to identify the language of the sentences.
The scores of those which cannot be identified as the corresponding language are reranked by a discount of $\alpha$. 
$\alpha$ is a hyperparameter.

The second reranking mechanism is based on \textbf{n-gram coverage}. 
Because the sentence pairs are scored independently, redundancy may exist in those high-score sentences.
To enhance the diversity of the selected corpus, we first sort the sentence pairs in the descending order based on their scores.
Next we maintain a $n$-gram pool for source sentences, and scan the source sentences from the top down.
Those sentences that have no $n$-gram different from those in the pool will receive a discount of $\beta$, and both $n$ and $\beta$ are hyperparameters. 

Note that before reranking, we always normalize the score according to their rankings, so that scores provided by different models can be unified. The score of the $i$-th sentence pair is:
\begin{equation}
    score_i = 1 - \frac{rank_i}{N}
\end{equation}
where $i$-th pair ranks $rank_i$ in all the sentence pairs and $N$ denotes total number of pairs.

%We also try to rerank the scores with the help of the language models so that the unfluent sentence pairs are punished. 
%However, the final performance does not seem to improve in our experiments.
We also try to rerank through language models, but it does not bring improvements.
Thus we do not use this reranking mechanism in our submissions.

\subsubsection{Ensemble}
Different models may capture different features during training and inference.
To make use of group wisdom and improve the final performance, we ensemble the following four models by averaging scores:
\begin{itemize}
    \item Model 1: Standard XLM + Linear Layer. The learning rate of XLM and linear layer are $1e^{-8}$ and $1e^{-5}$ respectively. 
    \item Model 2: Standard XLM + Linear Layer. The learning rate of XLM and linear layer are $1e^{-7}$ and $1e^{-4}$ respectively. 
    \item Model 3: Standard XLM + Convolutional Layer. The learning rate of XLM and linear layer are $5e^{-7}$ and $5e^{-4}$ respectively. 
    \item Model 4: Small XLM + Linear Layer. The learning rate of XLM and linear layer are $5e^{-7}$ and $5e^{-4}$ respectively.
\end{itemize}
All the models use $16$ batch size per GPU. 

\section{Experiments}

We conduct various experiments to evaluate the performance of different models, and select the most proper hyperparameters for both Khmer-English and Pashto-English. 
Note that \textbf{FS} and \textbf{FT} denote \textbf{From Sratch} and \textbf{Fine-Tune} respectively.

Firstly, we conduct the experiments with both the provided aligned sentence pairs (denoted as  \textbf{Baseline}) and our mined data at the first iteration of the mining module.
It shows that our system can outperform the baseline remarkably and the ensemble of four different models can further improve the performance. 
As Table~\ref{tab:experiment-1} illustrates, Model 1-4 outperform baseline by about $1\sim2$ BLEU in both km-en and ps-en . Besides, the ensemble model performs the best in general.

\begin{table}[htbp]
\centering
\caption{BLEU Scores of Difference Models}
\label{tab:experiment-1}
\begin{tabular}{lcccc}
\multirow{2}{*}{Model} & \multicolumn{2}{c}{km-en} & \multicolumn{2}{c}{ps-en}  \\ 
\cmidrule(lr){2-3} \cmidrule(lr){4-5}
  & FS & FT & FS & FT\\
\hline
Baseline & 7.28 & 10.24 & 9.81 & 11.37 \\
Model 1 & 8.33 & 11.43 & 11.21 & 13.11\\
Model 2 & 8.96 & 11.38 & \textbf{11.43} & 13.18 \\
Model 3 & 8.72 & 11.29 & 11.26 & 12.74\\
Model 4 & 9.01 & 11.27 & 11.36 & 13.09 \\
Ensemble & \textbf{9.22} & \textbf{11.51} & 11.28 & \textbf{13.52} \\
\hline
\end{tabular}
\end{table}

Next, to verify the effectiveness of the iterative mining strategy in the mining module, we compare the performance of the same ensemble model with different mined data.
In our paper, we iteratively mine data for three times, and combine them with the provided sentence-aligned corpus. Table~\ref{tab:experiment-2-data} reveals the mining scale each time.
As table~\ref{tab:experiment-2} shows, iteration 3 works best for km-en and iteration 2 for ps-en respectively.

\begin{table}[htbp]
\centering
\caption{The Number of Mined Sentence Pairs}
\label{tab:experiment-2-data}
\begin{tabular}{lcc}
Data & km-en & ps-en \\
\hline
Data 1 & 238K & 200K \\
Data 2 & 330K & 120K \\
Data 3 & 660K & 20K \\
\hline
\end{tabular}
\end{table}

\begin{table}[htbp]
\centering
\caption{BLEU Scores with Different Mined Data }
\label{tab:experiment-2}
\begin{tabular}{lcccc}
\multirow{2}{*}{Data} & \multicolumn{2}{c}{km-en} & \multicolumn{2}{c}{ps-en}  \\ 
\cmidrule(lr){2-3} \cmidrule(lr){4-5}
  & FS & FT & FS & FT\\
\hline
+ Data 1 & 9.22 & 11.51 & 11.28 & 13.52 \\
+ Data 1+2 & 9.47 & 11.56 & \textbf{12.17} & \textbf{13.19} \\
+ Data 1+2+3 & \textbf{9.84} & \textbf{11.62} & 12.14 & 12.69 \\
\hline
\end{tabular}
\end{table}

Finally, by introducing the reranking mechanism, we can further improve the performance, which is shown by Table~\ref{tab:experiment-3-km} and ~\ref{tab:experiment-3-ps}. 
Note that $\alpha=0$ or $\beta=0$ means it does not have any discount.
We select $\alpha=0.2,n=2,\beta=0.2$ and $\alpha=0,n=1,\beta=0.1$ for km-en and ps-en for our submissions.

\begin{table}[htbp]
\centering
\caption{BLEU Scores with Reranking for km-en}
\label{tab:experiment-3-km}
\begin{tabular}{lcc}
  & FS & FT \\
\hline
$\alpha=0,\beta=0$& 9.84 & 11.62 \\
$\alpha=0.2,n=2,\beta=0.05$& 10.40 & 12.25 \\
$\alpha=0.2,n=2,\beta=0.1$& 10.38 & 11.87 \\
$\alpha=0.2,n=2,\beta=0.2$& \textbf{10.50} & \textbf{12.45} \\
$\alpha=0.2,n=3,\beta=0.1$& 10.09 & 12.09 \\
$\alpha=0.3,n=2,\beta=0.05$& 10.40 & 12.25 \\
\hline
\end{tabular}
\end{table}

\begin{table}[htbp]
\centering
\caption{BLEU Scores with Reranking for ps-en}
\label{tab:experiment-3-ps}
\begin{tabular}{lcc}
  & FS & FT \\
\hline
$\alpha=0,\beta=0$& 12.17 & 13.19 \\
$\alpha=0,n=1,\beta=0.1$& \textbf{12.28} & 13.34 \\
$\alpha=0.2,n=1,\beta=0.1$& 12.15 & 13.06 \\
$\alpha=0,n=2,\beta=0.1$& 12.20 & \textbf{13.38} \\
$\alpha=0,n=2,\beta=0.2$& 12.20 & 13.31 \\
\hline
\end{tabular}
\end{table}

\section{Conclusion}

In this paper, we present our submissions to the WMT20 shared task on parallel Corpus filtering and alignment for low-resource conditions.
Our Volctrans system consists of two modules: a) Mining module is responsible for mining potential parallel sentence pairs out of the provided document pairs.
Word alignment model is utilized and an iterative mining strategy is further taken to boost the mining performance.
b) Scoring module aims to evaluate sentence pairs quality according to their parallelism and fluency properties, by exploiting an XLM-based scorer.
We further tune the output score with different reranking mechanism, by considering language detection confidence and n-gram vocabulary coverage. 
Finally, four models are ensembled to improve the final performance.
We also make some analysis through a variety of experiments.
% In the future, we plan to 
% %investigate how to accelerate the scoring procedure (e.g., knowledge distilling, pruning, etc.). Besides, we will 
% evaluate the performance when applying our systems to high-resource language pairs, e.g., English-German or English-French.

\bibliographystyle{acl_natbib}
\bibliography{emnlp2020}

% \appendix

% \section{Appendices}

% \section{Supplemental Material}

\end{document}